%% file: sdu-main.tex
\def\year{2022}\relax
\documentclass[letterpaper]{article} 
\usepackage{sdu/aaai22}  
\usepackage{times}  
\usepackage{helvet}  
\usepackage{courier}  
\usepackage[hyphens]{url}  
\usepackage{graphicx} 
\usepackage{natbib}  
\usepackage{caption} 
\DeclareCaptionStyle{ruled}{labelfont=normalfont,labelsep=colon,strut=off} 
\usepackage{algorithm}
\usepackage{algorithmic}

\usepackage{newfloat}
\usepackage{listings}

\usepackage{booktabs, multirow}
\usepackage{xcolor}
\usepackage{enumitem}
\usepackage{caption}
\usepackage{subcaption}

\floatstyle{ruled}
\newfloat{listing}{tb}{lst}{}
\floatname{listing}{Listing}
\title{Neural Architectures for Biological Inter-Sentence Relation Extraction}

\title{Neural Architectures for Biological Inter-Sentence Relation Extraction}
\author {
    Enrique Noriega-Atala,
    Peter M. Lovett, 
    Clayton T. Morrison,
    Mihai Surdeanu
}
\affiliations {
    The University of Arizona, Tucson, AZ, USA\\
    \{enoriega,plovett,claytonm,msurdeanu\}@email.arizona.edu
}

\begin{document}

\maketitle

\input{contents.tex}

\bibliography{acl/anthology,custom}

\end{document}

%% file: contents.tex
\newcommand\blue[1]{\textcolor{blue}{#1}}

\begin{abstract}
    We introduce a family of deep-learning architectures for \emph{inter-sentence} relation extraction, i.e.,  relations where the participants are not necessarily in the same sentence. We apply these architectures to an important use case in the biomedical domain: assigning biological context to biochemical events. In this work, biological context is defined as the 
type of biological system within which the biochemical event is observed.
The neural architectures encode and aggregate {\em multiple} occurrences of the same candidate context mentions to determine whether it is the correct context for a particular event mention. We propose two broad types of architectures: the first type 
aggregates multiple instances that correspond to the same candidate context with respect to
event mention before emitting a classification; the second type 
independently classifies each instance and uses the results to vote for the final class, akin to an ensemble approach. Our experiments show that the proposed neural classifiers are competitive and some achieve better performance than previous state of the art traditional machine learning methods without the need for 
feature engineering. Our analysis shows that the neural methods particularly improve precision compared to traditional machine learning classifiers and also demonstrates 
how the difficulty of inter-sentence relation extraction increases as the distance between the event and context mentions increase.
\end{abstract}
        
\section{Introduction}
\input{introduction.tex}

\section{Related Work}

\input{relatedwork.tex}

\section{Neural Architectures for Context Association 
}\label{sec:architecture}
\input{architecture.tex}

\section{Full-Text Context-Event Relation Corpus}
\input{dataset.tex}

\section{Experiments and Results}\label{sec:experiments}
\input{experiments.tex}


\section{Conclusions}
\input{conclusions.tex}

%% file: introduction.tex
\newcommand\evt[1]{\colorbox{pink}{\texttt{#1}}}
\newcommand\ctx[1]{\colorbox{cyan}{\texttt{#1}}}
\newcommand\auxmarker[1]{\colorbox{lightgray}{\texttt{#1}}}

Extracting biochemical interactions that describe mechanistic information from scientific literature is a task that has been well studied 
by the NLP community~\citep{cohen2015,zhou_zhong_he_2014,Hirschman2005}.
Automated event detection systems such as~\citep{Valenzuela2017,riedel2011model,kilicoglu2011adapting,quirk2011msr,bjorne2011generalizing,Bjrne2018BiomedicalEE,10.1093/bioinformatics/btaa540,rao2017biomedical} are able to detect and extract biochemical events with high throughput and good recall. The information extracted with such tools enables scientists and researchers to analyze, study and discover mechanistic pathways and their characteristics by aggregating the interactions and biological processes described in the scientific literature.

However, when dealing with such mechanistic processes it is important to identify the \emph{biological context} in which they hold. 
Here, biological context means the type of biological system, described at different levels of granularity, such as species, organ, tissue, cellular component, and/or cell-line within which the extracted biochemical interactions are observed.
Knowing the biological context is important to correctly interpret the mechanistic pathways described by the literature. For example, some tumors associated with oncogenic Ras in humans are different from those in mice, suggesting that the Ras pathway differs in both species~\citep{hamad2002distinct}. Ignoring the biological context information, specifically the species in the prior example, can mislead the reader to draw incorrect conclusions.
\begin{table}
    \centering
    \begin{tabular}[htb]{lc}
        \toprule
        \textbf{Quantity} & \textbf{Count} \\
        \midrule
        \# of inter-sent. relations & 1936 \\
        Mean sent. distance & 22 \\
        Median sent. distance & 5 \\
        Max sent. distance & 225 \\
        \bottomrule
    \end{tabular}
    \caption{Statistics about the inter-sentence distances of biological context annotations.}\label{tab:distances}
\end{table}


\begin{figure*}[htb]
    \scriptsize
    Transfection of the R-Ras siRNA effectively reduced the expression of endogenous R-Ras protein in \ctx{PC12 cells}.
 
    These results demonstrate that activation of endogenous R-Ras protein is essential for the ECM mediated cell migration and that regulation of R-Ras activity plays a key role in ECM mediated cell migration.
\evt{Sema4D and Plexin-B 1-Rnd1 inhibits PI3-K activity} through its R-Ras GAP activity.
\caption{Example of an inter-sentence relation annotated by a domain expert. The biological context, highlighted in blue, is established two sentences prior to the event mention, highlighted in pink.}\label{fig:intersentence}
\end{figure*}

\begin{figure}[tb]
    \centering
    \includegraphics[width=.35\textwidth]{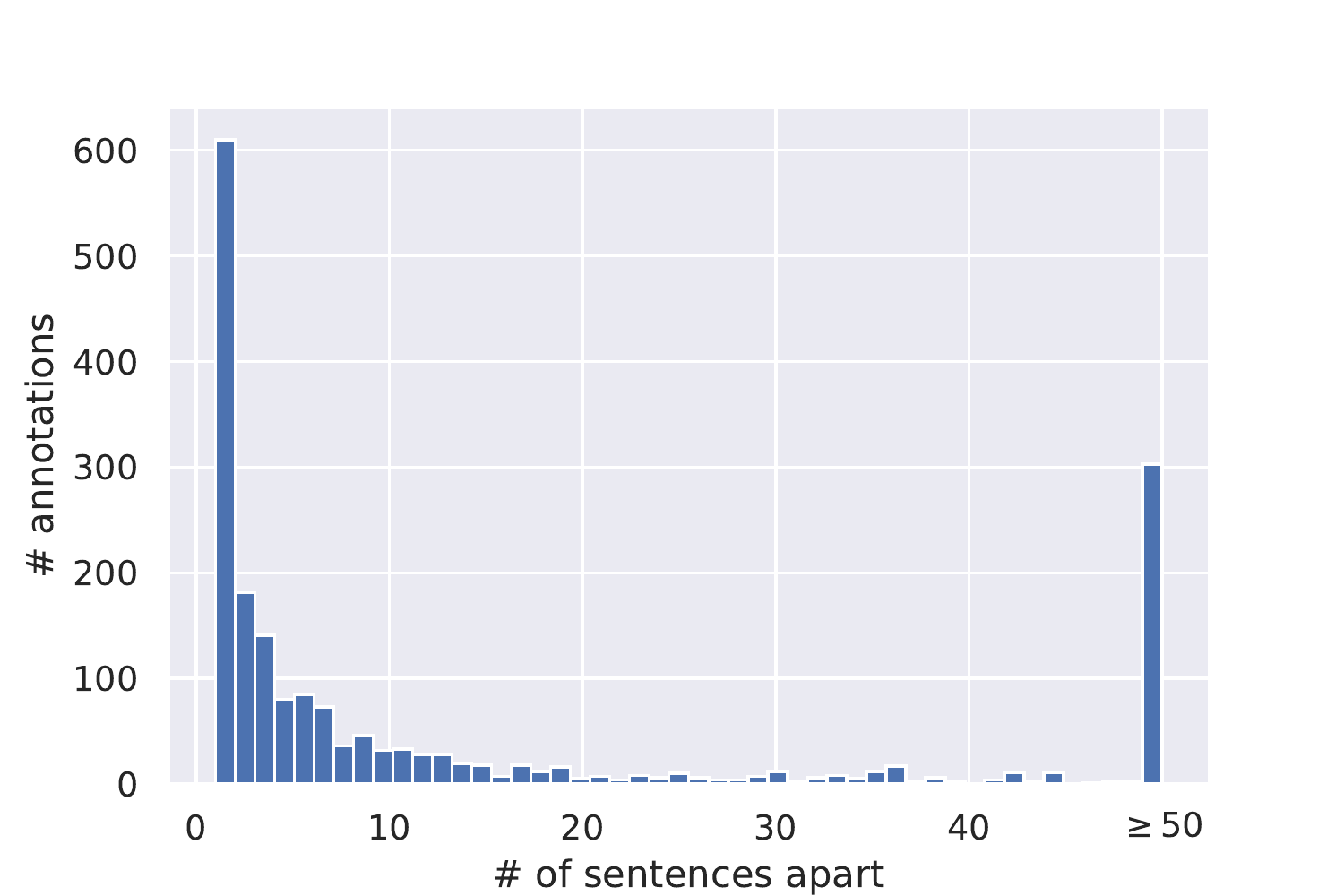}
    \caption{Distribution of inter-sentence distances of biological context annotations.}\label{fig:hist-distances}
\end{figure}

Biological context is often
not explicitly stated in the same clause that contains the biochemical event mention. Instead, 
the context is often established explicitly somewhere else in the text, such as the previous sentence or paragraph. In other words, there is a \emph{long distance relation} between the event mention and its context. In these cases, the context is implicitly propagated through the discourse that leads up to that particular biochemical event mention, as illustrated in figure~\ref{fig:intersentence}. Table \ref{tab:distances} and figure \ref{fig:hist-distances} contain summary statistics about the sentence distances for the relations in the corpus used in 
this work. These statistics indicate that, while most of the inter-sentence relations are close to the event mention they are associated with, there is a long tail of biological context mentions 
that are further than five sentences away from the corresponding event mentions.

We frame the problem of associating event mentions with their biological context as an inter-sentence relation extraction task and propose a family of deep-learning architectures to identify context. 
The approach inspects an event mention, a candidate context mention, and the text between them to determine whether the candidate context mention \emph{is context of} the event mention. Our work makes the following contributions:
\begin{itemize}
	\item Proposes a family of neural architectures that leverages large pre-trained language models for multi-sentence relation extraction.
    \item 
    Extends a corpus of cancer-related open access papers with biochemical event extractions annotated with biological context. Unlike the original corpus, this extended data set includes the full text of each article, tokenized and aligned to its annotations.
    \item 
    Analyzes multiple methods to aggregate different pieces of evidence that correspond to the same input event and context, and 
    assesses the overall performance and reliability of the networks under these different aggregation schemes.
\end{itemize}

%% file: relatedwork.tex
\begin{figure*}[tb]
	\scriptsize
	\begin{enumerate}[label=(\alph*)]
		\item Phospholipase C delta-4 overexpression upregulates  \evt{<EVT>} ErbB1/2 expression \evt{</EVT>} , Erk signaling pathway , and proliferation in \ctx{<CON>} MCF-7 \ctx{</CON>} cells .
		\item Phospholipase C delta-4 overexpression upregulates \evt{<EVT>} ErbB1/2 expression \evt{</EVT>} , Erk signaling pathway , \ldots have linked the upregulation of \auxmarker{[EVENT]} with rapid proliferation in certain \auxmarker{[CONTEXT]} \ldots \ctx{<CON>} MCF-7 \ctx{</CON>} cells .
		\item \ldots \ctx{<CON>} macrophages \ctx{</CON>}, and \auxmarker{[CONTEXT]}, where it is a trimeric complex consisting of one alpha-chain \ldots \auxmarker{[SEP]} \ldots FcRbeta also acts as a chaperone that increases \evt{<EVT>} FcepsilonRI expression \evt{</EVT>}
		
	\end{enumerate}
	\caption{Example input text spans. (a) Single-sentence segment with markers; (b) multi-sentence segment with markers and masked secondary event and context mentions; and (c) truncated long multi-sentence segment.}
    \label{fig:examples}
\end{figure*}

The problem of {\em relation extraction} (RE) has received extensive
attention \cite{banko2007,Bach2007}, including within the biomedical
domain \cite{Quan2014,Fundel2007}, with recent promising results
incorporating distant supervision \cite{poon2015}. However, most of the work focuses on identifying relations among entities within the
same sentence. In the biological context association problem, the entities are potentially located in different sentences, 
making the context association task an instance of 
an \emph{inter-sentence} relation extraction problem.

Previous work in inter-sentence relation extraction includes~\cite{swampillai2011},
which combined within-sentence syntactic features with an introduced dependency
link between the root nodes of parse trees from different sentences
that contain a given pair of entities. \citep{sahu-etal-2019-inter} proposes an inter-sentence relation extraction model that builds a labeled edge graph convolutional neural network model on a document-level graph. There have 
also been efforts to create language resources to foster the development of inter-sentence relation extraction methods. \citep{yao-etal-2019-docred} propose an open domain data set generated from Wikipedia to Wikidata. \citep{mandya-etal-2018-dataset} propose an inter-sentence relation extraction data set constructed using distance supervision. 
Modeling inter-sentence relation extraction using transformer architectures require processing potentially long sequences. Long 
input sequences are problematic because computing the self-attention matrix has quadratic runtime and space complexity relative to the its length. This observation has motivated research efforts to generate efficient approximations of self-attention.  \citep{Beltagy2020Longformer} proposes a sparse, drop-in replacement for the self-attention mechanism with linear complexity that relies on sliding windows and selects domain-dependent global attention tokens from the input sequence. \citep{wang2020linformer} proposes a lower-rank approximation of the self-attention matrix to linearize the complexity. \citep{tay2021synthesizer} ommits the pair-wise dependencies between the input tokens and then factorizes the attention matrix to reduce its rank. Other approaches \citep{choromanski2021rethinking} rely on kernel functions to compute approximations with linear time and space complexity. \cite{Chen2021PermuteFormerER} takes this approach further by using relative position encodings, instead of absolute ones.

Prior work has specifically studied
the contextualization of information extraction in the biomedical domain. \cite{gerner2010exploration} associates anatomical contextual containers with 
event  mentions that appear in the same sentence 
via a set of rules that considers lexical patterns in the case of ambiguity and falls back to token distance if no pattern is matched. \cite{sarafraz2012finding} elaborates on the same idea by incorporating dependency trees into the rules instead of lexical patterns, as well as introducing a method to detect negations and speculative statements.

\citep{noriega-atala2020} previously studied the task of context association for the biomedical domain and framed it as a problem of inter-sentence relation extraction. 
This work presents
set of linguistic and lexical features that describe 
the neighborhood of the participant entities and proposes 
an aggregation mechanism that results in improved context association.

Previous work relied upon feature engineering to encode the participants and their potential interactions. State-of-the-art NLP research 
leverages large language models to exploit transfer learning. Models such as~\citep{devlin-etal-2019-bert}, and similar transformer based architectures~\citep{vaswani2017attention} better capture the semantics of text based on its surrounding context with unsupervised pre-training over extremely large corpora. Specialized models, such as \cite{Liu2019RoBERTaAR,lee2019,alsentzer-etal-2019-publicly} refine language models by continuing pre-training with in-domain corpora. 


To the best of our knowledge, the work presented here 
is the first 
to propose and analyze deep-learning aggregation and ensemble architectures for many-to-one, long-distance relation extraction.

%% file: architecture.tex
\begin{figure*}[tb]
	\centering
    \includegraphics[width=\textwidth]{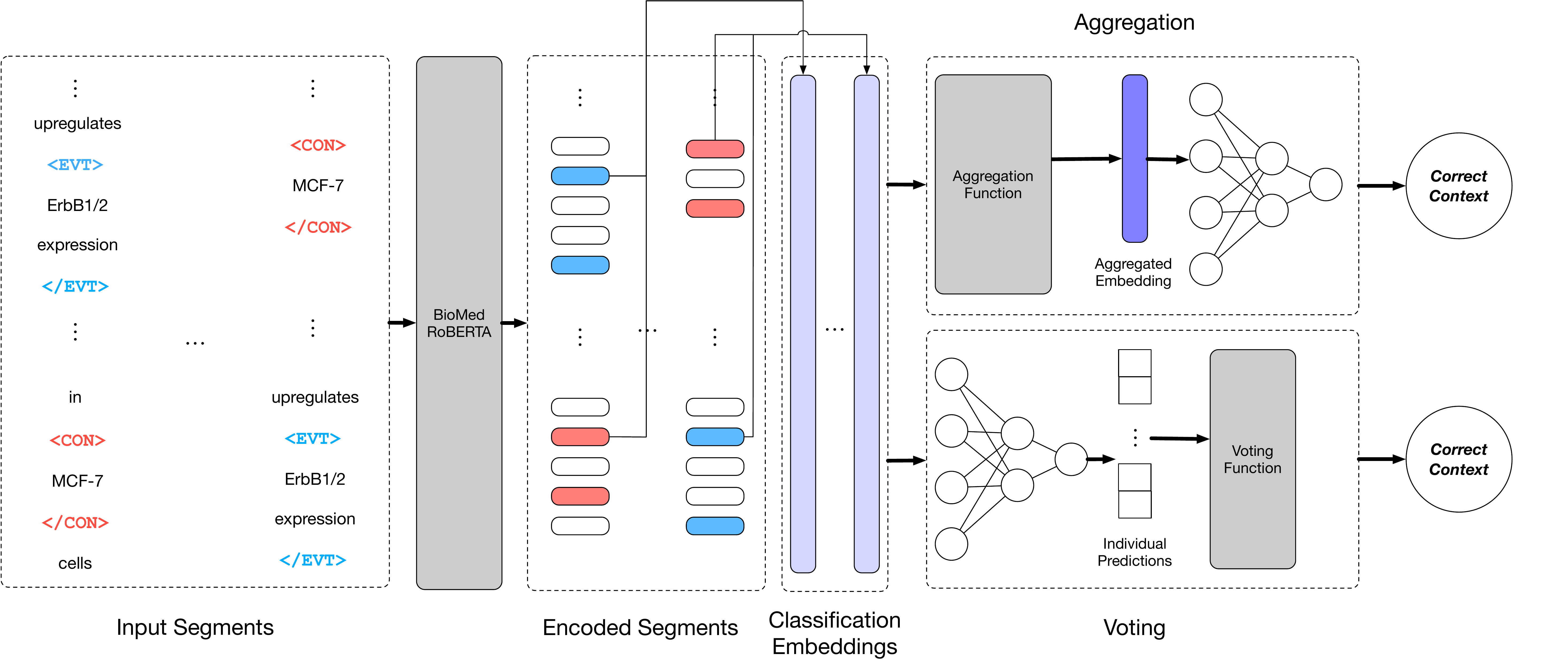}
    \caption{Context association neural architecture. The left-most box represents the input text segments after pre-processing. The blocks inside the encoded segments box represent BioMed RoBERTa's hidden states for the input segments. The classification embeddings box contains averages of the hidden states corresponding to the \texttt{<EVT>} and \texttt{<CON>} tokens of each input segment. Depending on the choice of architecture, classification embeddings either flow through (a) the aggregation block, which combines them to then generate the final classification; or (b) the voting block, where each embedding is classified, then the final result is generated through a voting function.}
    \label{fig:architecture}
\end{figure*}

We propose a family of 
neural architectures designed to determine whether a candidate \emph{context class} is relevant to a given biochemical event mention. A biochemical event mention (\emph{event mention} for short) describes the interaction between proteins, genes, and other gene products through biochemical reactions such as 
regulation, inhibition, phosphorylation, etc. In particular, we %
focus on the 12 interactions detected by REACH~\citep{reach2018}. A biological container context mention (\emph{context mention} 
for short) represents an instance from any 
of the following biological container types: 
species (e.g., human, mice), organ (e.g., liver, lung), tissue type (e.g., endothelium, muscle tissue), cell type (e.g., macrophages, neurons), or cell line (e.g., HeLa, MCF-7).




In this work, we use 
an existing information extraction 
system~\citep{Escarcega:2018} to detect and extract event mentions and candidate context mentions. Candidate context mentions are grounded 
to 
ontology concepts with 
unique identifiers to accommodate 
different spellings and synonyms that refer to the same biological container type. 
The specific ontology depends on the type of entity: 
UniProt\footnote{\url{https://www.uniprot.org/}} for proteins, PubChem\footnote{\url{https://pubchem.ncbi.nlm.nih.gov/}} for chemical entities, etc.


\begin{figure}
	\centering
	\includegraphics[width=0.35\textwidth]{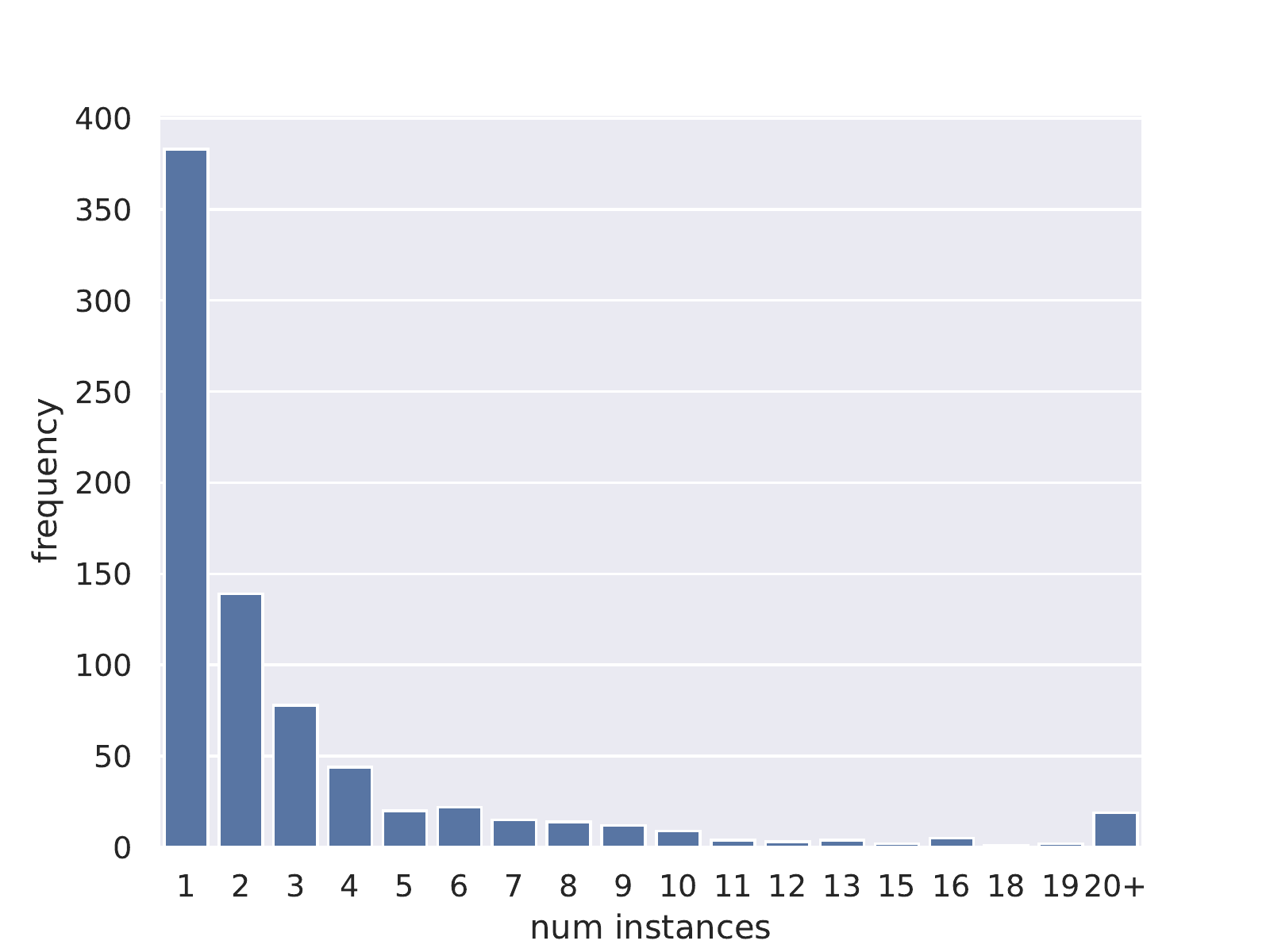}
	\caption{Distribution of the number of context class detections per article ($n_i$).}\label{fig:ni}
\end{figure}

Importantly, a context biological container type 
is likely mentioned multiple times in the document. Approximately half of the context container types 
in the 
context-event relation corpus are detected two or more times, as illustrated in figure \ref{fig:ni}.
Every candidate context mention that refers to the same container type 
is paired with the relevant event mention to generate a text segment for each pair. Each segment is represented as 
the concatenation of the sentences that include the event mention, one 
mention of the candidate context container type, 
and all the sentences in between. These text segments are used as input to the network to make predictions. If an article contains $n_i$ context mentions of container type 
$i$, then for each event mention the network will take up to $n_i$ input text segments to determine if type 
$i$ is a context of the event. The task of the network is to learn whether context type 
$i$ is a context of the 
specific event mention by looking at a subset of the $n_i$ inputs. An article with $j$ context types 
and $m$ event mentions will see a total of $j \times m$ classification problems and a total of $\sum_i^j n_i \times m$ input text segments. Figure~\ref{fig:architecture} shows a block diagram of the family of architectures.

Each input segment is preprocessed as follows. The boundaries of the relevant event and candidate context mentions are marked with the special tokens: \texttt{<EVT>...</EVT>} for the event mention and \texttt{<CTX>...} \texttt{</CTX>} for the context mention. Other event or context mentions present in the segment are masked with special \texttt{[EVENT]} or \texttt{[CONTEXT]} tokens, respectively, to avoid confusing the classifier with other event mentions that aren't the focus of the current prediction. Figure \ref{fig:examples} shows 
example 
text spans where the event 
and context mentions are surrounded by their boundary tokens.
Next, each preprocessed text segment is tokenized using the tokenizer specific to the pre-trained transformer used as the encoder.
If a tokenized sequence exceeds the maximum length allowed by the transformer, it is truncated before the encoding step by selecting the prefix of the sequence up to half the length, the suffix up to half the length minus one token, and inserting a special \texttt{<SEP>} token between them. Any truncated input segment is guaranteed to retain both mentions and their local lexical context. Figure \ref{fig:examples} shows an example of a  segment truncated using this procedure.
After tokenization, the segments are encoded using BioMed RoBERTa-base~\citep{domains}~\footnote{We used the available public checkpoint for both the BPE and BioMed RoBERTa models from \url{https://huggingface.co/allenai/biomed_roberta_base}}, based on~\citep{Liu2019RoBERTaAR}. 

The 
output hidden states of the \texttt{<EVT>} and \texttt{<CON>} tokens are averaged to create a \emph{classification embedding}.

Each classification task emits a single binary prediction, but has up to $n_i$ classification embeddings to account for the multiple (potential) context mentions that originate from the previously discussed process. To generate a single prediction, the network must combine the information carried forward by the classification embeddings. We propose two general approaches to combine the classification embeddings and generate the final prediction 
by combining the information {\em before} classification and {\em after} classification, respectively:

\begin{itemize}
	\item \textit{Aggregation}: Classification embeddings are combined together using an aggregation function. The aggregated embedding is then passed through a multi-layer perceptron (MLP) to emit a binary classification.
	\item 
	\textit{Voting}: Each classification embedding is passed individually through the MLP, which emits a local decision based only on the individual input text segment. The individual decisions are combined using a voting function to emit the final classification.
\end{itemize}

\begin{table*}[hbt]
	\centering
  \begin{tabular}{lcrrr}
	\toprule
	& \textbf{Documents} & \textbf{Event mentions} & \textbf{Context mentions} & \textbf{Annotations} \\
    \midrule
    Validation & 6 & 685 & 713 & 1,192\\
    Cross validation & 20 & 1,169 & 1,926 & 1,543\\
    Total & 26 & 1,854 & 2,639 & 2,735\\
    \midrule
    \multicolumn{5}{c}{\emph{Cross-validation split}}\\
    Training & 3 & 975.83 (58.32) & 1,654.83 (52.83) & 1,288.33 (95.89)\\
    Testing & 17 & 193.16 (58.32) & 271.16 (52.83) & 254.66 (95.89)\\
	\bottomrule
  \end{tabular}
  \caption{Statistics of the context association dataset. The upper part shows statistics from the overall dataset, both in total and split by the two partitions: (a) validation set, and (b) partition used for the formal cross-validation experiments. The lower part shows the average and standard deviations used for train/test for the different folds in cross-validation.}
  \label{tab:dataset-stats}
\end{table*}

Intuitively, aggregation functions consider multiple information points to make an informed decision based on the ``bigger picture'' presented by the article. Voting functions, on the other hand, make isolated decisions solely based on information local to each input text segment, then use those individual predictions to vote for the final classification, akin to an ensemble approach.

There are multiple ways to implement aggregation and voting functions. We propose four implementations of each kind, each following a intuitive principle.

\begin{figure*}[!tb]
	\centering
	\begin{subfigure}[b]{0.49\textwidth}
		\includegraphics[width=\textwidth]{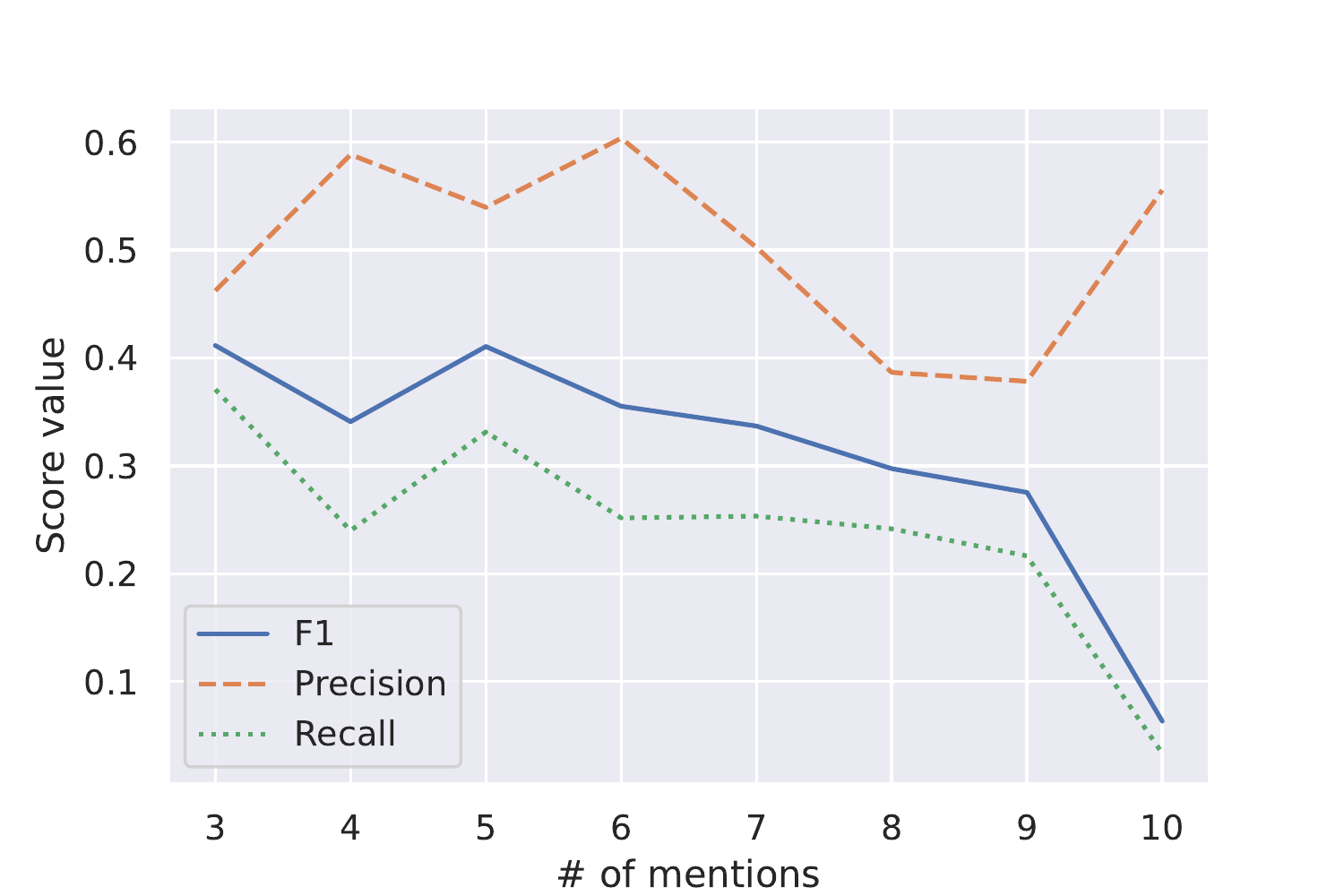}
		\caption{Majority vote}
	\end{subfigure}
	\hfill
	\begin{subfigure}[b]{0.49\textwidth}
		\includegraphics[width=\textwidth]{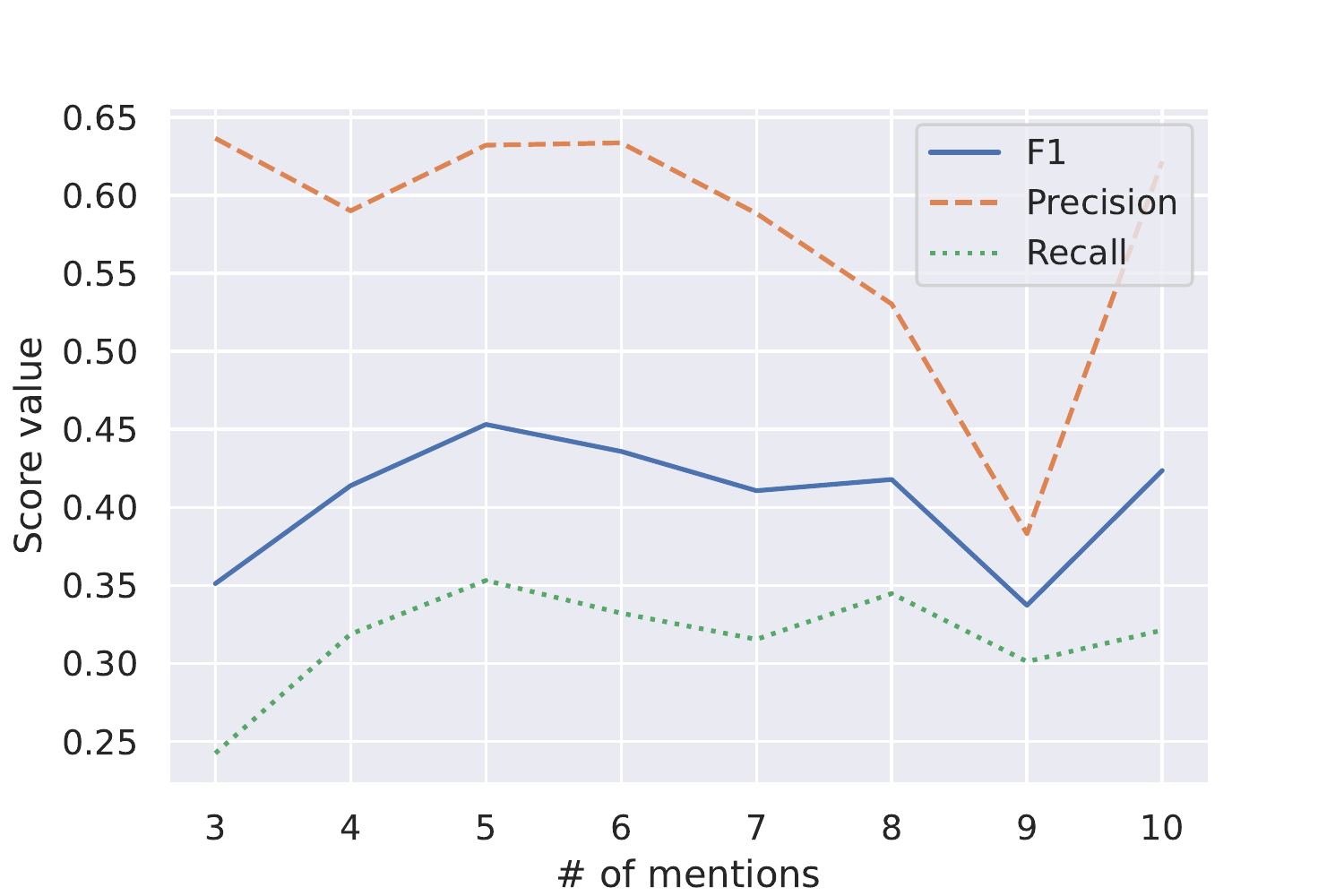}
		\caption{Average aggregation}
	\end{subfigure}

    \caption{Precision/recall/F1 scores of the relation classifier as the number of context mention considered for each individual relation classification is varied.}
    \label{fig:num-mentions}
\end{figure*}

\subsubsection*{Aggregation Functions}\label{sec:aggregations}


\emph{Nearest Context Mention}: Following the intuition that textual proximity should be a strong indicator of association, this approach selects the context mention of the relevant context type 
that is closest to the event mention. The closest context mention can appear either before or after event mention. In this setting, all other context mentions are ignored. The approach results in only one, unaltered classification embedding. It is equivalent to the case where only one mention of the relevant context type appears in a document ($n_i = 1$).

\emph{Average Context Embedding}: Conversely, all mentions of the candidate context type 
can bear a degree of responsibility to determine whether it is context of the event mention. Without making a statement about the importance of each context mention, we consider the text segments of the $k$ nearest context mentions of the relevant context type, to either side. The upper bound is enforced for efficiency and is left as a hyper parameter. If there are less than $k$ context mentions, all the text segments are considered. The segments are encoded, then the resulting classification embeddings are averaged.

\emph{Inverse Distance aggregation}: It can be argued that the influence of each context mention in the final decision decreases when it is farther 
apart from the event mention. We propose this aggregation approach, where instead of averaging the $k$ nearest classification embeddings, they are combined as a weighted sum, where each classification embedding's weight is defined as $w_i = d_i^{-1} / \sum_j^k d_j^{-1}$, the normalized inverse sentence distance between the event mention and the context mention. The resulting aggregated embedding still carries information from the nearest $k$ context mentions, but their contributions diminish inversely proportionally to their distance from the event mention.

\emph{Parameterized aggregation}: Instead of relying upon a heuristic approach to calculate the weights that determine the contributions of each classification embedding, we let the network learn the interactions between them using an attention mechanism. The parameterized aggregation approach concatenates $k$ nearest classification embeddings and uses a MLP to reduce the concatenated embeddings to a new vector with the same number of components as an individual classification embedding. The MLP works as map that combines the original $k$ classification embeddings whose parameters are learned during training. If the number of input text segments is $< k$, the concatenated classification embeddings are padded with zeros before being mapped to the new vector space. 

\subsubsection*{Voting Functions}


\emph{One hit}: This voting approach requires the minimum amount of evidence to trigger a positive classification. The context type 
is classified as \texttt{is context of} the event mention if \emph{at least one} classification embedding is classified as positive. Intuitively, this voting function favors 
recall.

\emph{Majority vote}: Conversely, it can be argued that there should be consensus in the vote. The majority vote function triggers a positive classification 
if at least half 
of the classification embeddings are classified as positive. In contrast to \emph{one hit}, this voting function favors 
precision.

\emph{Post-inverse distance vote}: Analogous to the inverse distance aggregation approach, 
this approach takes the vote of each classification embedding as 
weighted by the normalized inverse sentence distance: $w_i = d_i^{-1} / \sum_j^k d_j^{-1}$. The final classification is emitted in favor of the class with the highest weight. As opposed to the inverse distance aggregation approach, the combination happens \emph{after} passing the embeddings through the MLP.

\emph{Confidence vote}: We can weight each vote proportionally to the confidence of the classifier. In this approach, the vote of each individual classification is weighted by the classifier's confidence. The weights are given by the normalized logits of the vote of each classification embedding: $w_i = l_i / \sum_j^k l_j$.

%% file: dataset.tex

We used a corpus of biochemical events annotated with biological context to test the neural architectures for context assignment. 
Our version of the corpus is an extension of the corpus published by~\citep{noriega-atala2020}.

The corpus consists 
of automated extractions of 26 open-access articles from the PubMed Central repository, all related to the domain of cancer biology. The first type of extractions are \textit{events mentions}. An event mention is a relation between one or more entities participating in a biochemical reaction or its regulation. These mentions can be phosphorylation, ubiquitination, expression, etc. 
The second type of extractions are \textit{candidate context mentions}. These consist of 
named entity extractions of different biological container types: species, tissue types and cell lines.

Each event extracted was 
annotated by up to three biologists who assigned the event's 
relevant biological context from a 
pool of candidate context extractions available in the paper. Context annotations are not exclusive, meaning that every event mention can be annotated with one or more context classes. The result is a set of annotated events, where each event can have zero or more biological context associations, 
and there is at least one explicit mention for each biological context in the same article. 
The specifics of the automated event extraction procedure, annotation tool, annotations protocols and inter-annotator agreements are thoroughly detailed in~\citep{noriega-atala2020}. Table \ref{tab:dataset-stats} contains summary statistics of the data set's documents.

The original corpus release lacked 
the full text of the articles. Our proposed methodology requires the raw text to be used as input to the neural architectures. Our contribution here is an extension 
this corpus, 
where we identified, 
processed and tokenized the full text of the articles using the same information extraction tool~\cite{reach2018} used by the authors of the original corpus in such way that the tokens align correctly with the annotations and extractions published previously. The full-text context-event relation corpus, along with the code for the experiments presented in this document, is publicly available for reproducibility and further research.\footnote{\url{https://clulab.github.io/neuralbiocontext/}}

%% file: experiments.tex
In this section, we 
evaluate all proposed variants of the context association architecture and discuss the results. 

\subsection{Automatic Negative Examples}
The context-event relation corpus only contains positive 
context annotations of event mentions. 
We 
automatically generate negative examples for event mentions in each document by enumerating the cartesian product of all event 
and context mentions followed by 
subtracting the annotated pairs. 
One consequence of generating negative examples using 
this exhaustive strategy is that it results in most of the event/context pairs being 
negative examples, with 60,367 (95.68\%) negative pairs and 2,703 (4.32\%) positive pairs. This results in a severe class imbalance, which makes the classification task harder.


\subsection{Results and Discussion}

\begin{table}[!tp]
	\small
    \centering
    \begin{tabular}{lcccc}\toprule
        \textbf{Method}  &\textbf{Precision} &\textbf{Recall} &\textbf{F1} \\\midrule
    Majority (3 votes) &\textbf{0.580}* &0.498 &\textbf{0.536}* \\
    Parameterized agg. &0.537* &0.494 &0.514* \\
    One-hit &0.409 &0.668* &0.507 \\
    Post inv. distance &0.571* &0.446 &0.501 \\
    Nearest mention &0.541* &0.464 &0.499 \\
    Average (5 segs) &0.527 &0.469 &0.497 \\
    Inverse distance &0.544* &0.454 &0.495 \\
    Confidence vote &0.394 &0.443 &0.417 \\
    \midrule
	\multicolumn{4}{c}{\textit{Baselines}}&\vspace{.3em}\\
    Random forest &0.439 &0.541 &0.485 \\
    Logistic regression &0.361 &\textbf{0.699} &0.476 \\
    Heuristic &0.421 &0.548 &0.476 \\
    Decision tree &0.311 &0.389 &0.345 \\
    \bottomrule
    \end{tabular}
    \caption{Cross-validation 
    results for the \emph{is context of} class. 
    * denotes statistically significant improvement w.r.t. the random forest classifier.}\label{tab:results}
\end{table}

We use a cross validation evaluation framework 
similar to the evaluation methodology used by~\citep{noriega-atala2020}. Each fold contains all of the event-context 
pairs that belong to three different articles. 
However, we held out six papers as a development set.
During cross validation, one fold is used for testing and training is performed using the remaining $k-1$ folds plus the data from the development set. This way, we take advantage of more training data and avoid leaking the information from development into testing.

To better understand the impact of considering multiple context mentions at the time of aggregation or voting, we tuned this hyper parameter on the development set. Figure~\ref{fig:num-mentions} shows the effect of increasing the number of context mentions 
used for relation classification. The number of context mentions 
considered ranged from three to ten. 
Both architectures reach a peak F1 score between 3 to 5 context mentions. 
Performance quickly decays 
almost asymptotically, as the number of considered context mentions increases. This observation suggestions 
that increasing the number of 
input text segments derived from context mentions that are further apart from the event introduces too much noise into the decision process.

After the above tuning, we ran cross-validation experiments for all aggregation and voting methods. Based on the tuning results, we used the closest five mentions of each context class for the average aggregation architecture, and the closest three for all of the other 
architectures. Table~\ref{tab:results} summarizes
the cross validation performance scores for 
all the architecture variants. 
The precision, recall, and F1 scores reported are computed just for the positive class (i.e., {\em is context of}) to avoid artificially inflating the scores with the dominating negative class. 

The top performing architecture is the majority vote. It achieves an 
F1 score slightly above $0.53$. The majority vote architecture trades off recall for precision. The reason for this is that the architecture needs to 
see at least half of the individual input segments classified as positive in order to make 
that prediction. As a result, a positive classification using this architecture comes with a relatively high confidence.  As expected, the one-hit architecture achieves the opposite: it trades 
precision for recall. One-hit only needs to see one individual positive classification in order to emit a positive final classification. As a result, one-hit attains the highest recall within the neural architectures but is more prone to false positives.

We include several baseline algorithms to compare the performance of the neural architectures. The first baseline is a ``heuristic'' method that associates all the context types within a constant number of sentences to 
an event mention. We also include our implementation of three classifiers using the feature engineering method of~\citep{noriega-atala2020}. The top three performing neural architectures have statistically significantly higher F1 score than the random forest classifier, which is the strongest baseline algorithm.

Note that the methods proposed by~\citep{noriega-atala2020} that are 
included in the table {\em aggregate multiple feature vectors} from the different context mentions 
into a new feature vector composed of multiple statistics from 
the original feature space. 
Examples of these feature aggregations include 
the minimum, maximum and average values of the distribution of sentence distances, the frequency of the context type, and 
the proportion of times the context mention is part of a noun phrase.
Their aggregation approach is analogous to the one presented here 
(although here 
we operate in embedding space), which is why the comparison between these two approaches 
is fair. 


Table~\ref{tab:by-distance} lists the classification scores of the top performing method, stratifying the data by the sentence distance to the closest context mention of the relevant class. Performance, along with 
the frequency of such instances, quickly degrades as the distance between event and context mention increases.

\begin{table}
    \small
\centering
\begin{tabular}{ccccc}
    \toprule
    \textbf{Distance} &  \textbf{Precision} &  \textbf{Recall} &     \textbf{F1} &  \textbf{Support} \\
    \midrule
    0                    &      0.796 &   0.818 &  0.807 &    573 \\
1                    &      0.490 &   0.450 &  0.469 &    262 \\
2                    &      0.398 &   0.336 &  0.364 &    146 \\
3                    &      0.531 &   0.402 &  0.457 &    107 \\
4                    &      0.569 &   0.393 &  0.465 &     84 \\
5+                   &      0.214 &   0.131 &  0.163 &    351 \\
    \bottomrule
\end{tabular}
\caption{Cross-validation scores for the positive class of the Majority (3 votes) architecture stratified by sentence distance to the closet context mention of the same class.}\label{tab:by-distance}
\end{table}

%% file: conclusions.tex
We propose a family of neural architectures to detect biological context of biochemical events. We approach the problem as an \emph{inter-sentence} relation extraction that uses multiple pieces of document-level evidence to classify whether a specific context label 
is the correct context type of an 
event extraction.

We provide an analysis of different methods to combine evidence to generate a final decision. The approaches work either before classification, by aggregating embeddings in order to emit a decision, 
or after classification, creating ensembles that vote for multiple individual decisions.

Using an expert-annotated corpus that associates biochemical events with relevant biological context, our results show that in spite of the severe class imbalance, several the neural architectures are competitive and achieve higher classification performance than a deterministic heuristic and other machine learning approaches. 

The neural architectures particularly favor precision, which makes them more appealing for applications 
where higher precision is desirable. 

Inter-sentence relation extraction continues to be a challenge. An ablation study of the degree of 
aggregation of evidence shows how considering mentions that are further apart from the event degrades 
performance. An error analysis by sentence distance shows how the difficulty of inter-sentence relation extraction correlates with the distance between the participants. The result of these analyses suggest that understanding how to filter out noisy event-context mention pairs and how to better weight the contribution of long-spanning mention pairs are important directions for future research.